\documentclass{article}

\usepackage{arxiv}

\usepackage[utf8]{inputenc} 
\usepackage[T1]{fontenc}    
\usepackage{hyperref}       
\usepackage{url}            
\usepackage{booktabs}       
\usepackage{amsfonts}       
\usepackage{nicefrac}       
\usepackage{microtype}      
\usepackage{lipsum}
\usepackage{graphicx}

\usepackage[numbers]{natbib}

\usepackage{hyperref}
\hypersetup{
    colorlinks=true,
    linkcolor=black
}

\usepackage{tabularx}
\usepackage{graphicx} 
\usepackage{amsmath} 
\usepackage{amssymb} 
\usepackage{amssymb} 
\usepackage{subcaption} 
\usepackage{booktabs}
\usepackage{multirow}
\usepackage{makecell} 
\usepackage[most]{tcolorbox}   
\usepackage{tcolorbox}
\usepackage{threeparttable}

\usepackage[framemethod=default]{mdframed}
\newmdenv[
  topline=false,
  bottomline=false,
  rightline=false,
  leftline=true,
  linecolor=gray,
  linewidth=2pt,
  backgroundcolor=gray!5,
  innertopmargin=0.8\baselineskip,
  innerbottommargin=0.8\baselineskip,
  innerleftmargin=1em,
  skipabove=1em,
  skipbelow=1em
]{PromptBox}
\usepackage{enumitem}
\usepackage{float}

\newif\ifmarkchanges     
\markchangesfalse          

\usepackage{xcolor}

\ifmarkchanges
  \newcommand{\ins}[1]{\textcolor{red}{#1}}
  \newcommand{\del}[1]{\textcolor{blue}{\sout{#1}}}
  \usepackage[normalem]{ulem} 
\else
  \newcommand{\ins}[1]{#1}
  \newcommand{\del}[1]{}     
\fi

\graphicspath{ {./images/} }

\title{MolProphecy: Bridging Medicinal Chemists' Knowledge and Molecular Pre-Trained Models via a Multi-Modal Framework}

\usepackage{authblk}

\newcommand{\correspondingauthorA}{\textsuperscript{*}}
\newcommand{\correspondingauthorB}{\textsuperscript{\textdagger}} %

\author[1]{Jianping Zhao}
\author[1]{Qiong Zhou}
\author[4]{Tian Wang}
\author[3]{Yusi Fan}
\author[3]{Qian Yang}
\author[6]{Li Jiao}
\author[5]{Chang Liu}
\author[3]{Zhehao Guo}
\author[3]{Qi Lu}
\author[2,3]{\normalsize Fengfeng Zhou\correspondingauthorA} 
\author[2,3]{\normalsize Ruochi Zhang\correspondingauthorB} 

\affil[1]{College of Computer Science and Technology, Changchun University of Science and Technology, Changchun, Jilin, 130012, China}
\affil[2]{College of Computer Science and Technology, Jilin University, Changchun, Jilin, 130012, China}
\affil[3]{Key Laboratory of Symbolic Computation and Knowledge Engineering, Ministry of Education, Jilin University, Changchun, Jilin, 130012, China}
\affil[4]{Department of Chemical and Biological Engineering, The Hong Kong University of Science and Technology, Clear Water Bay, Hong Kong, 999077, China}
\affil[5]{Beijing Life Science Academy, Beijing, 102209, China}
\affil[6]{Communication University of China, Beijing, 102209, China}

\affil{\correspondingauthorA Corresponding author: \texttt{FengfengZhou@gmail.com}}
\affil{\correspondingauthorB Corresponding author: \texttt{zrc720@gmail.com}}

\begin{document}
\maketitle
\begin{abstract}
MolProphecy is a human-in-the-loop (HITL) multi-modal framework designed to integrate chemists’ domain knowledge into molecular property prediction models. While molecular pre-trained models have enabled significant gains in predictive accuracy, they often fail to capture the tacit, interpretive reasoning central to expert-driven molecular design. To address this, MolProphecy employs ChatGPT as a virtual chemist to simulate expert-level reasoning and decision-making. The generated chemist knowledge is embedded by the large language model (LLM) as a dedicated knowledge representation and then fused with graph-based molecular features through a gated cross-attention mechanism, enabling joint reasoning over human-derived and structural features. Evaluated on four benchmark datasets (FreeSolv, BACE, SIDER, and ClinTox), MolProphecy outperforms state-of-the-art (SOTA) models, achieving a 15.0\% reduction in RMSE on FreeSolv and a 5.39\% improvement in AUROC on BACE. Analysis reveals that chemist knowledge and structural features provide complementary contributions, improving both accuracy and interpretability. MolProphecy offers a practical and generalizable approach for collaborative drug discovery, with the flexibility to incorporate real chemist input in place of the current simulated proxy—without the need for model retraining. The implementation is publicly available at \url{https://github.com/zhangruochi/MolProphecy}.
\end{abstract}

\section{Introduction}

Drug discovery is a complex and resource-intensive process requiring deep expertise across chemistry, biology, and pharmacology. Medicinal chemists play a central role in designing molecules that maximize biological efficacy while minimizing side effects. Despite their efforts, the drug development pipeline remains slow and heavily dependent on trial-and-error approaches, largely due to the difficulty of accurately predicting molecular properties and behaviors in complex biological systems. The process can take 10--15 years and cost between 1 and 2 billion dollars per approved drug \citep{dimasi2016innovation}. Moreover, despite technological advances, the cost and duration of drug development have steadily increased, a trend commonly referred to as Eroom’s law~\citep{scannell2012erooms}.

In recent years, artificial intelligence (AI) has shown substantial promise in accelerating this process, particularly through LLMs—from foundational architectures like BERT~\citep{devlin2019bert} and GPT-3~\citep{brown2020language} to subsequent influential models such as LLaMA~\citep{touvron2023llama} and Gemini~\citep{team2023gemini}—and domain-specific molecular pre-trained models. While AI-driven approaches offer faster and more scalable molecular property prediction, effectively integrating the qualitative, experience-based knowledge of medicinal chemists into quantitative AI models remains a major challenge. This tacit understanding, which includes mechanistic reasoning, stability considerations, and the effects of specific substituents, is not easily encoded in conventional data-driven models, which consequently struggle to capture the nuanced reasoning that underpins expert human judgment.

To address this challenge, we present MolProphecy, a HITL framework that couples chemist knowledge with structured molecular representations. Our approach concurrently encodes expert reasoning insights (currently derived from an LLM proxy) and extracts graph-based molecular features. These two distinct modalities are then dynamically fused via a gated cross-attention mechanism. This innovative integration enables MolProphecy to produce more accurate, robust, and interpretable predictions by aligning the strengths of both human-derived insights and structural information. The main contributions of this work are as follows:
\begin{itemize}
\item HITL integration. MolProphecy integrates chemist expertise as an independent knowledge modality: currently simulated via ChatGPT, yet seamlessly swappable with real chemist input without requiring model retraining.
\item Gated cross-attention fusion. A multi-modal architecture leverages a gated cross-attention mechanism to effectively align LLM-encoded chemist knowledge with GNN-derived structural features.
\item Empirical performance and interpretability. Empirical evaluations on MoleculeNet benchmarks demonstrate MolProphecy's superior performance over prior methods, accompanied by interpretable predictions facilitated by model visualization techniques.
\end{itemize}

\section{Related Work}

\subsection{Molecular Representation Models}
Pre-trained molecular representation models, which learn transferable chemical features from large-scale unlabeled data, have become central to computational chemistry by mitigating data scarcity and enhancing generalization. These models primarily follow two paradigms based on the input data type: sequence-based and structure-based. Sequence-based models, such as ChemBERTa\ \citep{chithrananda2020chemberta} and Mol-BERT\ \citep{li2021mol}, adapt Transformer architectures from natural language processing to SMILES strings, capturing chemical syntax and semantics. In parallel, structure-based approaches leverage Graph Neural Networks (GNNs), including foundational architectures such as Graph Isomorphism Network (GIN)\ \citep{xu2019powerful}, to learn embeddings directly from the molecular graph and capture richer topological information.

\subsection{Molecular Property Prediction Models}

Improving molecular property prediction, a fundamental task in drug discovery, has motivated work on a range of data modalities. Early studies compared different molecular representations and their predictive value \citep{yang2019analyzing}. Methods such as Neural Message Passing \citep{gilmer2017neural} and Graph Attention Networks (GAT)\citep{velickovic2018graph} advanced graph‐based learning, and large‐scale pre-training approaches like GROVER \citep{GROVER2020} and Graphormer \citep{Graphormer2021} used Transformer architectures to learn powerful structure-aware embeddings.

Recent research has turned to multi-modal frameworks. MolXPT \citep{liu2023molxpt} wraps molecules with textual context during generative pre-training, whereas SYN-FUSION \citep{sai2023synergistic} combines structural graphs with chemistry-guided information. MulAFNet \citep{ci2025mulafnet} integrates complementary molecular views through adaptive fusion, and MolPROP \citep{rollins2024molprop} unifies language and molecular graphs via joint representation learning.

Although these fusion models improve accuracy, they remain predominantly data-driven and provide limited chemical interpretability. This limitation highlights the need for frameworks that integrate qualitative domain expertise \citep{chen2023information}.

\subsection{LLMs and HITL in Drug Discovery}
LLMs have recently emerged as powerful tools in chemistry, with platforms like ChemCrow\ \citep{bran2024chemcrow} and Coscientist\ \citep{boiko2023autonomous} demonstrating their capability to assist in complex tasks like synthesis planning. Notably, studies show that models like GPT-4 can match or exceed the competencies of mid-level professional chemists on standardized exams and advanced reasoning benchmarks\ \citep{lin2025gpt4_taiwan_pharmacist, kim2024gpt4_kple}, suggesting their potential to significantly augment chemical research\ \citep{almeida2024investigating, jchemed2024_chatgpt_reactions}.

However, most systems use LLMs as external agents rather than structurally integrating their knowledge into predictive models. While HITL frameworks have been employed to align model behavior with expert preferences~\citep{nahal2024human, mosqueira2023human}, they rarely formalize domain knowledge within the model architecture itself. Recent work further shows that although LLMs are insufficient as standalone validators, combining them with automated methods and selective human oversight can achieve human-level quality with reduced manual effort~\citep{tsaneva2025knowledge}. To address this gap, our work proposes a cross-modal framework that explicitly fuses LLM-derived chemist insights with molecular representations. This approach, mirroring trends in high-stakes clinical applications such as FairCare~\citep{wang2024faircare}, aims to build more robust, interpretable, and expert-aligned predictive models for drug discovery.

\section{Method}
\subsection{Research Objectives}
\ins{Current molecular property predictors excel at pattern recognition yet lack the qualitative, experience-driven reasoning of medicinal chemists. Bridging this gap is fundamentally a HITL challenge: how can tacit chemist heuristics be injected into data-centric models without sacrificing scalability?}

\ins{MolProphecy addresses this gap with a multi-modal HITL framework that pairs graph-based molecular embeddings with chemist knowledge captured by an LLM proxy. We evaluate how this fusion improves predictive accuracy and interpretability, and we validate its transferability on four representative MoleculeNet datasets. Ultimately, we aim to provide a scalable template for expert-in-the-loop drug discovery workflows.}

\subsection{Overall Architecture of MolProphecy}
\begin{figure}[htbp]
    \centering
    \includegraphics[width=0.95\textwidth]{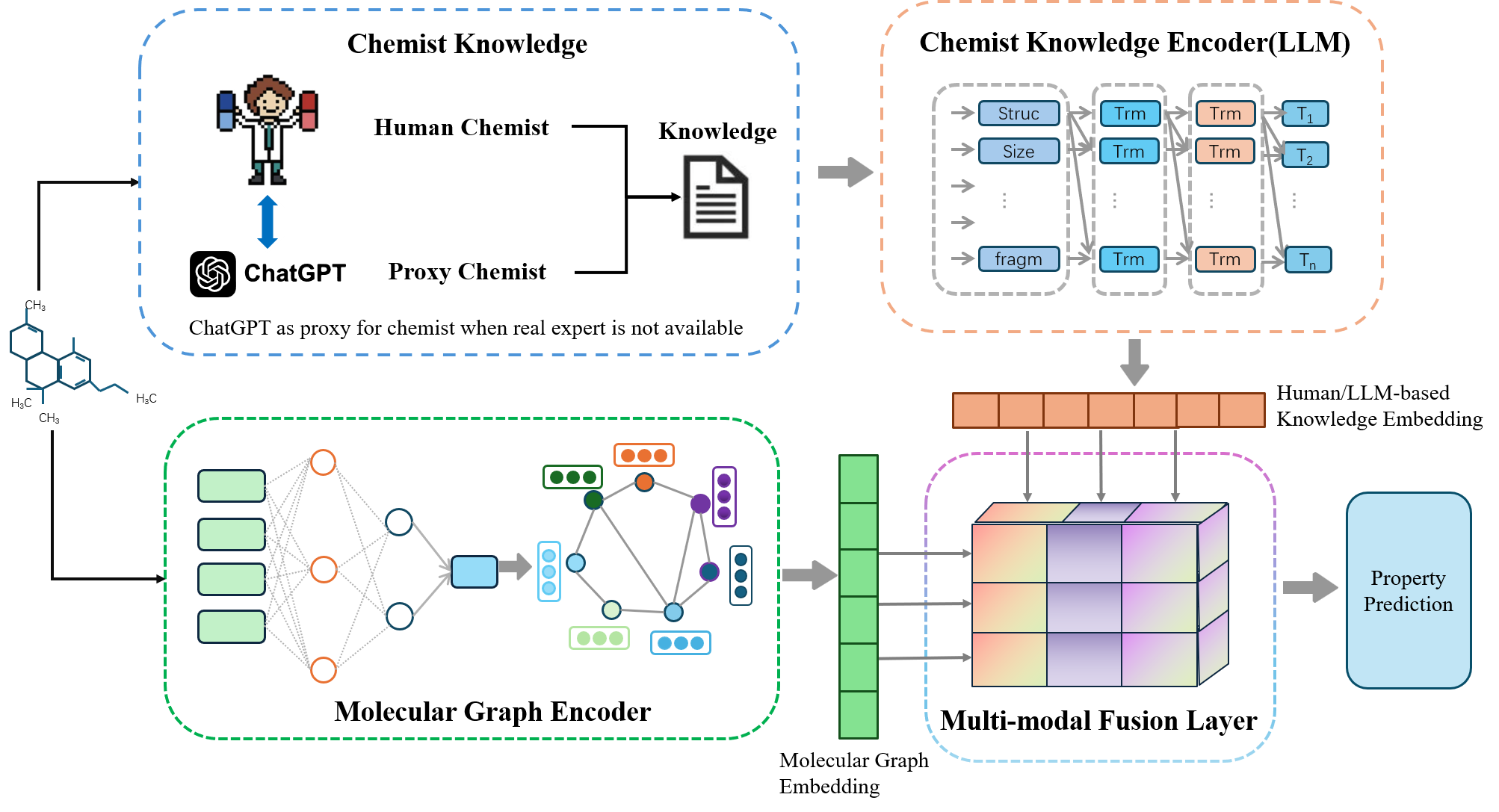}
    \caption{
        Overview of the proposed HITL framework for molecular property prediction. Chemist knowledge (collected from experts or generated by ChatGPT as a virtual proxy) and molecular structure features are encoded and fused via a multi-modal layer for downstream tasks. Trm refers to Transformer blocks within the LLaMA3-based Knowledge Encoder.
        }
    \label{fig:framework}
\end{figure}

Motivated by the need to incorporate chemist reasoning, MolProphecy integrates domain knowledge and structural representations via a multi-modal design. As illustrated in Figure \ref{fig:framework}, it comprises four components: (1) a chemist-knowledge generation module that collects knowledge insights from human chemist or ChatGPT; (2) a chemist-knowledge pathway that encodes these insights with an LLM; (3) a molecular-structure pathway that extracts graph-based features from molecular inputs; and (4) a multi-modal fusion module that integrates the two modalities through cross-attention.

The model is trained end-to-end for downstream property-prediction tasks. We next describe how chemist-like knowledge is simulated and embedded, forming one of the two core modalities in our model.

\subsection{Chemist Knowledge Simulation via ChatGPT}

A core innovation of our framework is the explicit integration of medicinal chemist expertise as a distinct information modality. As directly obtaining large-scale, consistently annotated insights from human chemists is often impractical, we employ ChatGPT as a virtual proxy to simulate their reasoning and commentary. By engineering a structured prompt, we systematically elicit domain-specific knowledge tailored to each molecule, creating a reproducible and scalable source of expert-level annotations.

Notably, it is designed to be agnostic to the knowledge source. In practical applications, real chemists can provide descriptions using the same structured template, ensuring compatibility and seamless integration into HITL workflows without model retraining.

\begin{figure}[htbp]
\centering
\includegraphics[width=0.95\textwidth]{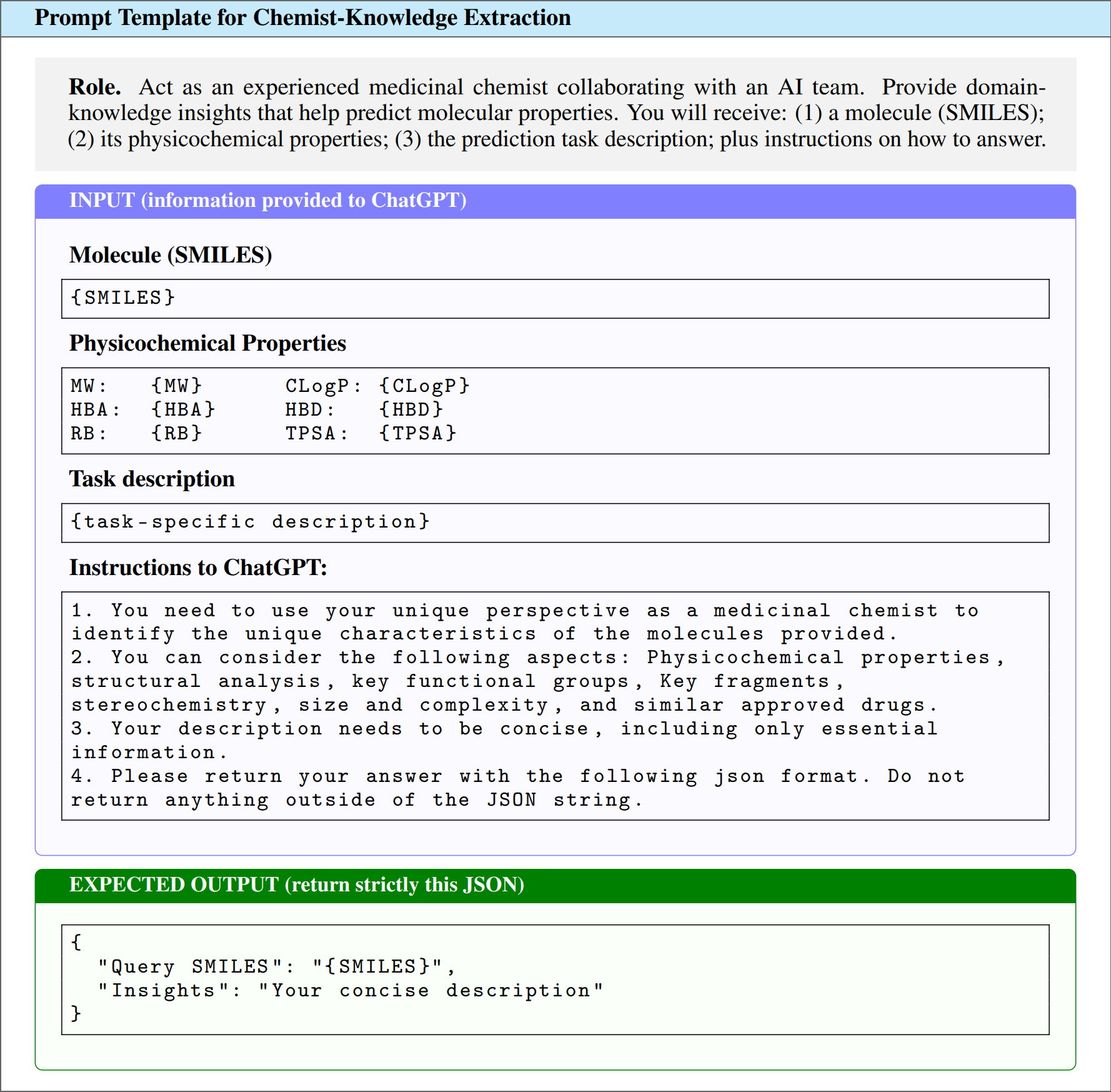}
\caption{Overall layout of the prompt template used for chemist-knowledge extraction. Full task-specific prediction objectives are provided in~\ref{task_descriptions}}
\label{fig:prompt_template}
\end{figure}

To ensure the extracted knowledge is contextually meaningful and actionable, our specialized prompt encompasses essential molecular context. This includes the structure (SMILES representation), computed physicochemical properties, a clearly stated predictive task, and explicit instructions guiding ChatGPT to emulate chemist reasoning. The complete prompt template is as follows in Figure~\ref{fig:prompt_template}. 

\subsection{Chemist Knowledge Encoder}
The Chemist Knowledge Encoder processes the reasoning insights generated by ChatGPT, leveraging an LLM to create rich, contextualized embeddings. Crucially, this module is designed to capture and represent high-level chemical reasoning, such as structural interpretations, functional group identification, physicochemical property inference, and ADMET risk assessment. This approach moves beyond merely treating the input as generic text, ensuring that the deep domain knowledge, central to our HITL approach, is effectively encoded.

The final chemist knowledge embedding \( \mathbf{h}_\text{chem} \in \mathbb{R}^{n \times d} \) is derived from the hidden states of the 13th transformer layer in the LLaMA3 model, where \( n \) is the sequence length and \( d \) is the embedding dimension. Prior work~\citep{chen2020generative} has shown that intermediate layers in generative LLMs often retain richer semantic representations than final output layers, particularly for transfer learning tasks. Formally:

\begin{equation}
    \mathbf{h}_\text{chem} = \text{LLaMA3}_{\text{layer}=12}(X),
\end{equation}

where \( X \) is the tokenized chemist insight text. The resulting sequence of token embeddings is passed to the cross-modal fusion module without additional pooling, allowing subsequent layers to learn optimal integration strategies.

\subsection{Molecular Graph Encoder}

The Molecular Graph Encoder employs a GIN to extract structural features from a molecular graph \( G = (V, E) \), where \( V \) denotes the set of atoms and \( E \) represents the set of chemical bonds. Each atom \( v \in V \) is initially associated with a feature vector \( \mathbf{h}_v^{(0)} \), capturing atomic descriptors such as element type, degree, and formal charge.

The GIN model updates node representations via iterative message passing over \( K \) layers. At the \( k \)-th iteration (\( k = 1, \dots, K \)), the node embedding \( \mathbf{h}_v^{(k)} \) is computed as:

\begin{equation}
    \mathbf{h}_v^{(k)} = \text{MLP} \left( (1 + \epsilon) \cdot \mathbf{h}_v^{(k-1)} + \sum\nolimits_{u \in \mathcal{N}(v)} \mathbf{h}_u^{(k-1)} \right),
\end{equation}

where \( \epsilon \) is a learnable scalar parameter, and \( \mathcal{N}(v) \) denotes the set of neighbors of node \( v \). This formulation aggregates information from local neighborhoods and integrates it with the previous state of each node. After \( K \) iterations, a graph-level molecular representation \( \mathbf{h}_\text{mol} \) is obtained via a READOUT function that aggregates node embeddings:

\begin{equation}
    \mathbf{h}_\text{mol} = \text{READOUT}(\{\mathbf{h}_v^{(K)} \mid v \in V\}).
\end{equation}

In our implementation, we adopt mean pooling as the READOUT function:

\begin{equation}
    \mathbf{h}_\text{mol} = \frac{1}{|V|} \sum_{v \in V} \mathbf{h}_v^{(K)}.
\end{equation}

\subsection{Multi-modal Fusion Layer}
\begin{figure}[htbp]
\centering
\includegraphics[width=0.95\textwidth]{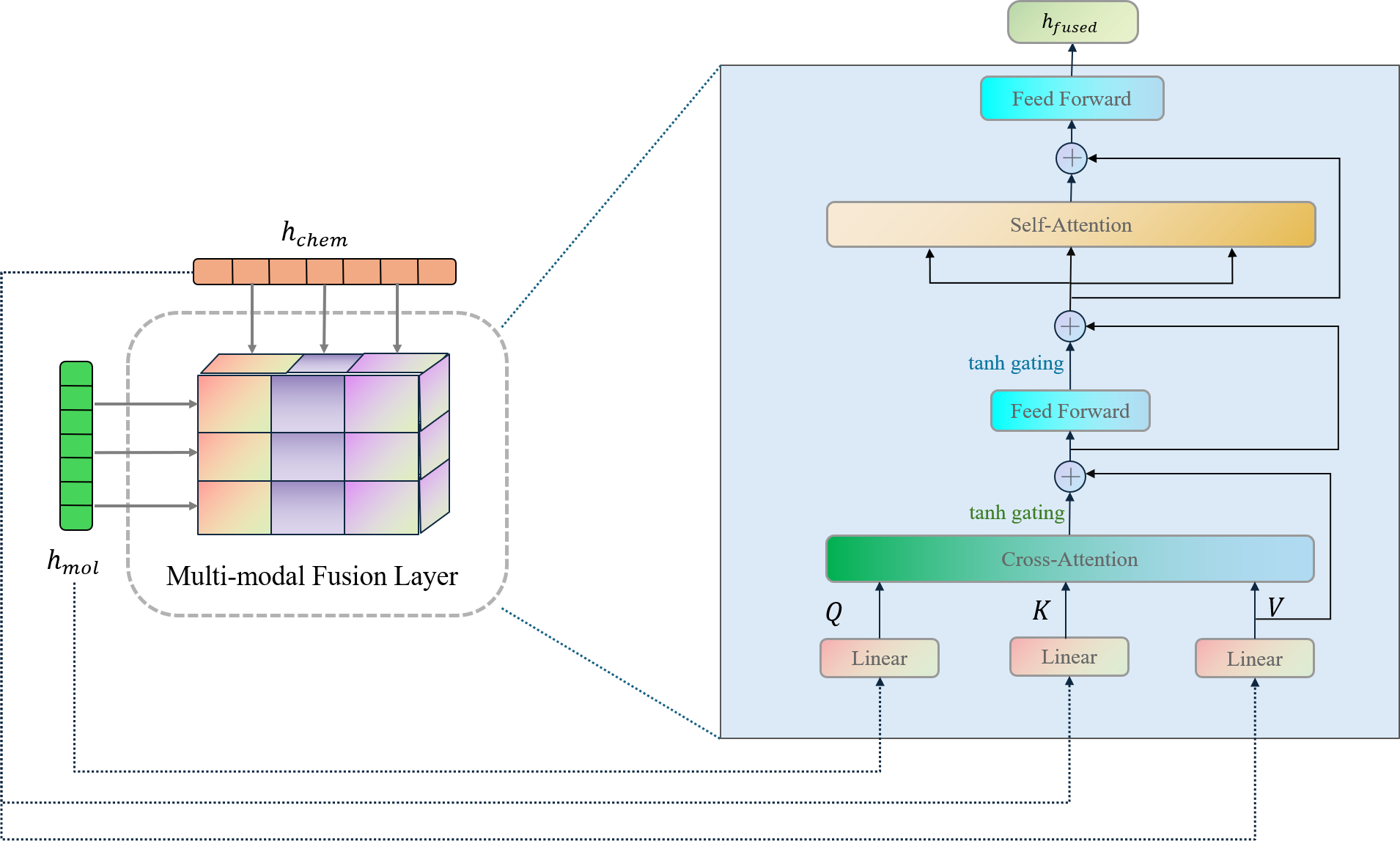}
\caption{Architecture of the Multi-modal Fusion Layer. It illustrates the flow from molecular and chemist knowledge embeddings through multi-head cross-attention, gated feedforward, self-attention, and feedforward layers to produce the final fused representation.}
\label{fig:gated_xattn_layer}
\end{figure}

The multi-modal fusion layer, illustrated in Figure~\ref{fig:gated_xattn_layer}, employs a gated multi-head cross-attention mechanism. In this setup, the molecular graph embedding acts as the query, while the chemist knowledge embedding serves as the key and value. This configuration reflects our domain intuition that primary structural features should be modulated and contextualized by the chemist knowledge insights. The learnable gating functions are intrinsically part of this mechanism, regulating the cross-modal information flow, and the design is inspired by recent advances in cross-modal attention for molecular applications\ \citep{liu2023molca}.

Let \( \mathbf{h}_\text{mol} \in \mathbb{R}^{n \times d} \) denote the molecular graph embedding and \( \mathbf{h}_\text{chem} \in \mathbb{R}^{m \times d} \) denote the chemist knowledge embedding, where \( n \) and \( m \) are the number of molecular and chemist knowledge tokens respectively, and \( d \) is the embedding dimension.

\begin{equation}
    \mathbf{A}_\text{cross} = \text{MultiheadAttention}(\mathbf{h}_\text{mol}, \mathbf{h}_\text{chem}, \mathbf{h}_\text{chem}, \text{mask} = M_\text{chem}),
\end{equation}

where \( M_\text{chem} \) is a padding mask for the chemist knowledge input. The attended representation is modulated via a learnable gating parameter \( \alpha_{\text{xattn}} \), yielding an updated molecular representation:

\begin{equation}
    \mathbf{h}_\text{mol} \leftarrow \mathbf{h}_\text{mol} + \tanh(\alpha_{\text{xattn}}) \cdot \mathbf{A}_\text{cross}.
\end{equation}

Following the cross-attention step, a gated feedforward layer is applied:

\begin{equation}
    \mathbf{h}_\text{mol} \leftarrow \mathbf{h}_\text{mol} + \tanh(\alpha_{\text{dense}}) \cdot \text{FFN}(\mathbf{h}_\text{mol}),
\end{equation}

where FFN denotes a feedforward neural network with ReLU activation. To enhance the contextual understanding within the molecular representation, a self-attention module is then applied:

\begin{equation}
    \mathbf{A}_\text{self} = \text{SelfAttn}(\mathbf{h}_\text{mol}, \mathbf{h}_\text{mol}, \mathbf{h}_\text{mol}, \text{mask} = M_\text{mol}),
\end{equation}

resulting in a further refined embedding:

\begin{equation}
    \mathbf{h}_\text{mol} \leftarrow \mathbf{h}_\text{mol} + \mathbf{A}_\text{self}.
\end{equation}

Lastly, an additional feedforward layer produces the final fused representation:

\begin{equation}
    \mathbf{h}_\text{fused} = \mathbf{h}_\text{mol} + \text{FFN}(\mathbf{h}_\text{mol}),
\end{equation}

which integrates both structural and chemist knowledge features.

\subsection{Prediction and Training Strategy}

The fused representation \( \mathbf{h}_\text{fused} \) is fed into a multi-layer perceptron (MLP) for downstream molecular property prediction. The configuration of the output layer and the choice of loss function are determined by the specific task type.

For classification tasks, the output is a probability score computed using a sigmoid function:

\begin{equation}
    \hat{y} = \sigma(\mathbf{W} \mathbf{h}_\text{fused} + \mathbf{b}),
\end{equation}

with binary cross-entropy (BCE) loss:

\begin{equation}
    \mathcal{L}_{\text{pred}} = - \frac{1}{N} \sum_{i=1}^{N} \left[ y_i \log \hat{y}_i + (1 - y_i) \log (1 - \hat{y}_i) \right].
\end{equation}

For regression tasks, a linear projection produces continuous outputs:

\begin{equation}
    \hat{y} = \mathbf{W} \mathbf{h}_\text{fused} + \mathbf{b},
\end{equation}

with mean squared error (MSE) loss:

\begin{equation}
    \mathcal{L}_{\text{pred}} = \frac{1}{N} \sum_{i=1}^{N} (y_i - \hat{y}_i)^2.
\end{equation}

The model is trained end-to-end using only the prediction loss \( \mathcal{L}_{\text{pred}} \). During this process, the LLaMA3-based Chemist Knowledge Encoder serves as a fixed feature extractor with its parameters frozen. The backpropagation thus jointly optimizes the remaining learnable components—the Molecular Graph Encoder, the Multi-modal Fusion Layer, and the prediction head—ensuring that both modalities effectively contribute to the final task-specific representations.

\section{Experiments}
\subsection{Experimental Setup}

To rigorously evaluate the effectiveness of our proposed multi-modal fusion approach, we conduct experiments on a diverse set of molecular property prediction tasks, encompassing both classification and regression objectives. 

\paragraph{Environment} Experiments were run in PyTorch on an NVIDIA RTX 4090. Models were trained with Adam (initial $3\times10^{-4}$) and a plateau scheduler (factor 0.5, patience 5) for at most 100 epochs with early stopping. We used a batch size of 128, dropout 0.5 in the final MLP, and weight decay $10^{-3}$.

\subsection{Datasets}
\ins{We evaluated MolProphecy on four benchmark datasets from MoleculeNet~\citep{wu2018moleculenet}, selected to cover a range of chemical tasks. These include the FreeSolv dataset for hydration-free-energy regression, and three classification benchmarks: BACE for $\beta$-secretase inhibition (positive rate: $\sim$26\%), SIDER across 27 adverse-reaction subtasks with severe class imbalance, and ClinTox for two highly imbalanced toxicity endpoints (<10\% positive rate). Key statistics for these datasets are detailed in Table~\ref{tab:dataset}.}

\begin{table}[htbp]
\centering
\small
\caption{Descriptions of the datasets used in this study.}
\label{tab:dataset}
\begin{tabular}{llccl}
\toprule
\textbf{Category} & \textbf{Dataset} & \textbf{\# Molecules} & \textbf{\# Tasks} & \textbf{Task Type} \\
\midrule
Physical Chemistry & FreeSolv & 642 & 1 & Regression \\
Biophysics & BACE & 1513 & 1 & Classification \\
\multirow{2}{*}{Physiology} 
  & SIDER & 1427 & 27 & Classification \\
  & ClinTox & 1478 & 2 & Classification \\
\bottomrule
\end{tabular}
\end{table}

\begin{figure}[htbp]
    \centering
    \begin{subfigure}[b]{0.85\textwidth}
        \centering
        \includegraphics[width=\textwidth, height=0.35\textheight, keepaspectratio]{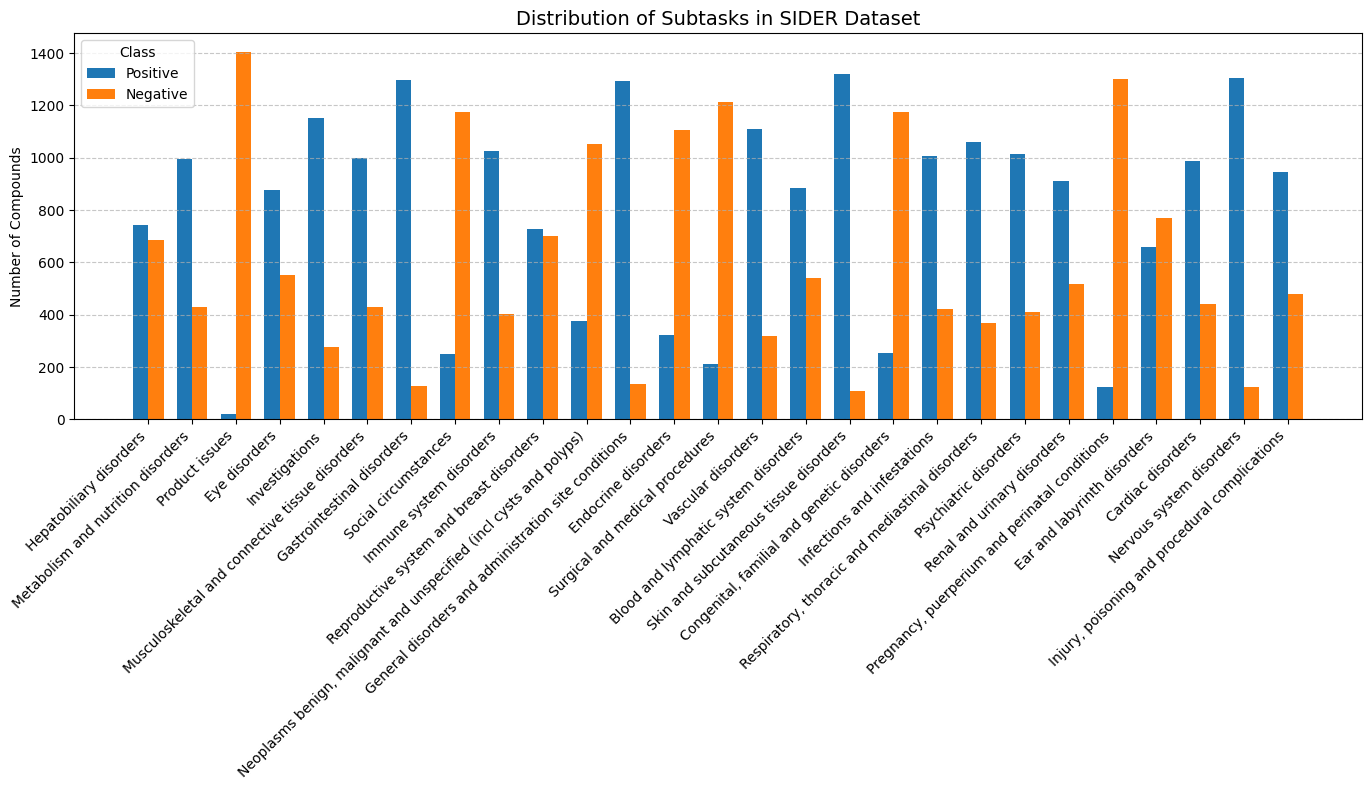}
        \caption{Distribution of positive and negative samples in the 27 SIDER adverse-reaction tasks.}
        \label{fig:sider-dist}
    \end{subfigure}
    \begin{subfigure}[b]{0.85\textwidth}
        \centering
        \includegraphics[width=\textwidth, height=0.25\textheight, keepaspectratio]{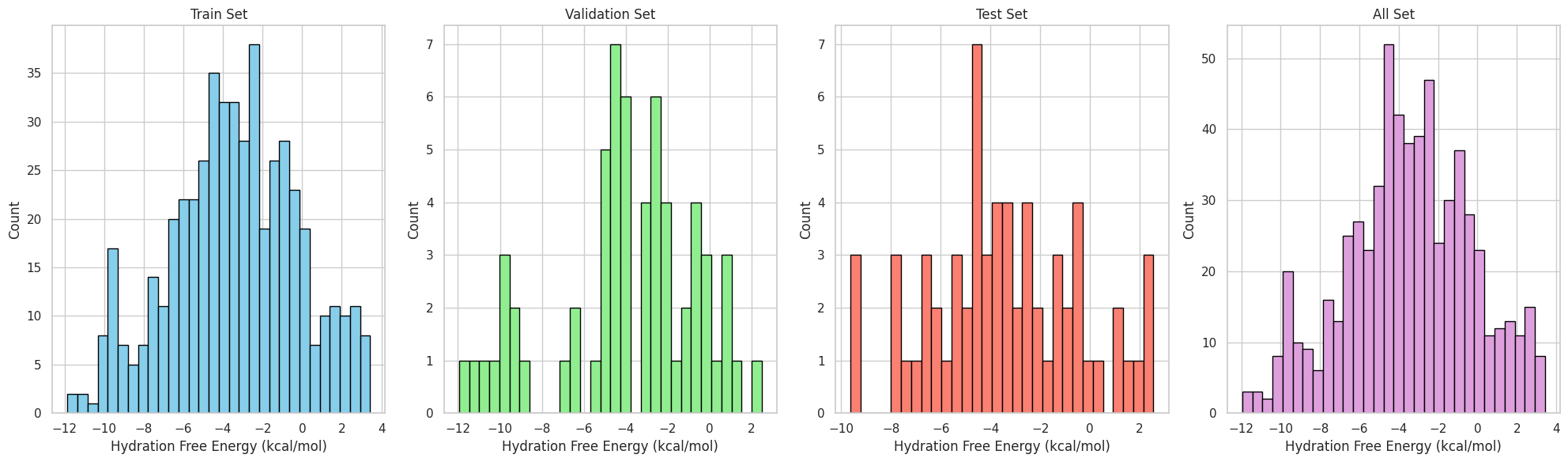}
        \caption{Distribution of FreeSolv hydration free energy values across different data splits.}
        \label{fig:freesolv-dist}
    \end{subfigure}
    \caption{Dataset characteristics motivating experimental design: (a) Task imbalance in SIDER and (b) split-wise distribution of FreeSolv hydration free energy values.}
    \label{fig:dataset-dist}
\end{figure}

\ins{For a robust evaluation of generalization, the classification datasets (BACE, SIDER, and ClinTox) were partitioned using the Bemis-Murcko scaffold splitting methodology~\citep{bemis1996scaffold}, a technique that prevents data leakage between sets. The smaller FreeSolv dataset, in contrast, employed a random split, with its target values standardized to mitigate the effect of outliers. The inherent challenges of these benchmarks are illustrated in Figure~\ref{fig:dataset-dist}. These include the extreme label imbalance across SIDER subtasks (Figure~\ref{fig:dataset-dist} (a)) and the broad, outlier-rich distribution of FreeSolv's target values (Figure~\ref{fig:dataset-dist} (b)), characteristics that necessitate models with high robustness to data imbalance and numerical instability.}

\subsection{Evaluation Metrics}

We evaluate performance using task-specific metrics. For classification tasks, we employ the Area Under the Receiver Operating Characteristic Curve (AUC-ROC), which measures the model's ability to discriminate between positive and negative classes:
\begin{equation}
\text{AUC-ROC} = \int_{0}^{1} \text{TPR}(\text{FPR}) \, d(\text{FPR})
\end{equation}
where $\text{TPR} = \frac{\text{TP}}{\text{TP} + \text{FN}}$ is the True Positive Rate and $\text{FPR} = \frac{\text{FP}}{\text{FP} + \text{TN}}$ is the False Positive Rate.

For regression tasks, our performance metric is the Root Mean Squared Error (RMSE), where lower RMSE indicates better predictive accuracy:
\begin{equation}
\text{RMSE} = \sqrt{\frac{1}{n} \sum_{i=1}^n (y_i - \hat{y}_i)^2},
\end{equation}
where $y_i$ represents the true value, $\hat{y}_i$ represents the predicted value, and $n$ is the number of samples.

This unified benchmarking setup enables a comprehensive and fair comparison across methods and tasks, supporting the robustness and generalizability of our proposed fusion framework.

\subsection{Performance Comparison with Baseline Models}
We evaluated the performance of MolProphecy by comparing it against a comprehensive set of SOTA baseline models spanning multiple modeling paradigms. These include classical graph neural networks such as GAT, SMILES-based language models like ChemBERTa, and large-scale pre-trained models including GROVER, Graphormer, and MolCLR. We also considered several recent fusion and geometry-aware approaches, such as MolHGT, SYN-FUSION, MolPROP, and MulAFNet. Together, these baselines represent a diverse and competitive benchmark that reflects current trends in molecular property prediction.

MolProphecy consistently outperforms all baseline models across the four MoleculeNet datasets, as summarized in Table~\ref{tab:molprophecy_comparison}. Notably, on FreeSolv, MolProphecy achieves an RMSE of 0.796, a 15.0\% reduction compared to the best baseline, SYN-FUSION (0.937). On SIDER, it attains an AUROC of 0.709, improving upon SYN-FUSION’s 0.699 by 1.43\%. For the BACE and ClinTox datasets, MolProphecy also demonstrates improved performance, with respective AUROC increases of 5.39\% and 1.06\% over previous SOTA models. These results underscore the effectiveness of the proposed multi-modal fusion strategy for integrating chemist knowledge with molecular structure.

\begin{table}[H]
\centering
\small
\caption{Performance comparison of different models across datasets. Bold values indicate the best performance. “–” denotes that the result was not reported in the original paper.}
\label{tab:molprophecy_comparison}
\begin{tabularx}{\textwidth}{lXXXX}
\toprule
\textbf{Method} & 
\textbf{FreeSolv (RMSE ↓)} & 
\textbf{BACE (ROC-AUC ↑)} & 
\textbf{SIDER (ROC-AUC ↑)} & 
\textbf{ClinTox (ROC-AUC ↑)} \\
\midrule

GAT\ \citep{velickovic2018graph} & 1.150 & 0.806 & 0.638 & 0.832 \\
ChemBERTa\ \parbox[t]{2.4cm}{\centering \citep{chithrananda2020chemberta}} & 0.994 & 0.857 & 0.680 & 0.888 \\
GROVER\ \citep{GROVER2020} & 1.99 & - & 0.658 & 0.944 \\
Graphormer\ \citep{Graphormer2021} & 2.09 & - & 0.620 & 0.881 \\
MolHGT\ \citep{deng2022describe} & 1.075 & 0.849 & 0.676 & 0.864 \\
MolCLR\ \citep{wang2022molecular} & 2.20 & 0.890 & 0.680 & 0.932 \\
SYN-FUSION\ \parbox[t]{2.4cm}{\centering \citep{sai2023synergistic}} & 0.937 & 0.805 & 0.699 & 0.947 \\
MolPROP\ \citep{rollins2024molprop} & 1.70 & 0.890 & – & 0.932 \\
MulAFNet\ \citep{ci2025mulafnet} & – & 0.875 & 0.636 & 0.939 \\

\textbf{MolProphecy (Ours)} & \textbf{0.796} & \textbf{0.938} & \textbf{0.709} & \textbf{0.957} \\
\bottomrule
\end{tabularx}
\end{table}

\subsection{Ablation Studies}
We conducted an ablation study to quantify the contribution of each modality, with results reported in Table~\ref{tab:ablation}. This study demonstrates that both modalities are crucial, as removing either leads to a significant drop in performance. Specifically, a model relying exclusively on the Chemist Knowledge Encoder (MolProphecy-Chem) fails to capture fine-grained structural details such as stereochemistry, whereas a model using only the GIN-based Molecular Graph Encoder (MolProphecy-GIN) lacks essential semantic cues from domain expertise. These findings confirm that our framework’s strength lies in fusing complementary information streams, specifically global functional insight from domain expertise and precise local geometry from the graph, to unlock its full predictive potential.

\begin{table}[htbp]
\centering
\small
\caption{
Ablation study results on four benchmark datasets. For the regression task (FreeSolv), we report RMSE (↓) under a random split. For classification tasks (BACE, SIDER, ClinTox), we report ROC-AUC (↑) under scaffold splits. MolProphecy-Chem uses only the Chemist Knowledge Encoder, while MolProphecy-GIN uses only the Molecular Graph Encoder. Bold values denote the best performance on each dataset.
}
\label{tab:ablation}
\begin{tabularx}{\textwidth}{lXXX}
\toprule
\textbf{Dataset} & \textbf{MolProphecy-Chem} & \textbf{MolProphecy-GIN} & \textbf{MolProphecy (Full)} \\
\midrule
FreeSolv (RMSE ↓) & 1.695 & 0.968 & \textbf{0.796} \\
BACE (AUROC ↑)    & 0.597 & 0.861 & \textbf{0.938} \\
SIDER (AUROC ↑)   & 0.501 & 0.631 & \textbf{0.709} \\
ClinTox (AUROC ↑) & 0.781 & 0.799 & \textbf{0.957} \\
\bottomrule
\end{tabularx}

\end{table}

\subsection{Feature Exploration and Visualization}
\begin{figure}[htbp]
    \centering
    \begin{subfigure}[b]{0.9\textwidth}
        \centering
        \includegraphics[width=\textwidth]{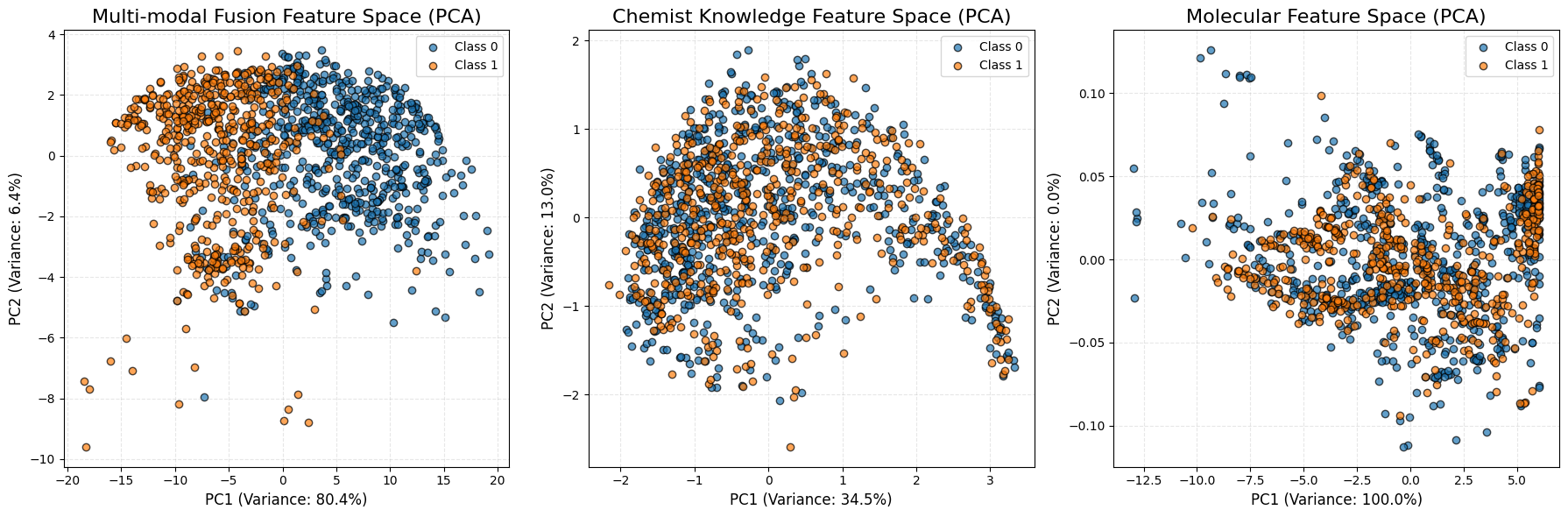}
        \caption{PCA projections of the learned representations on the BACE dataset: 
        (left) multi-modal fusion, (middle) chemist knowledge, and (right) molecular features. 
        Variance explained by PC1 and PC2 is indicated in each plot. 
        The fused space demonstrates better class separation and variance balance than either modality alone.}
    \end{subfigure}

    \vspace{1mm}

    \begin{subfigure}[b]{0.9\textwidth}
        \centering
        \includegraphics[width=\textwidth]{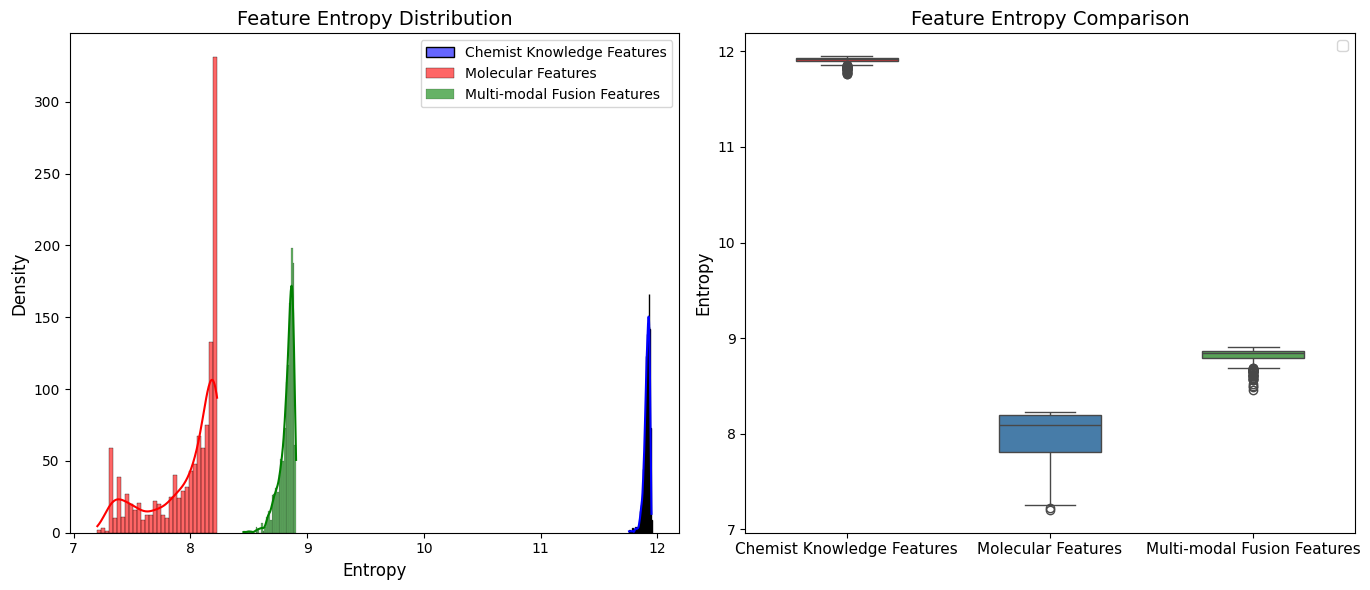}
        \caption{Feature entropy distribution and comparison on the BACE dataset.
        (Left) Density distribution of entropy values computed from chemist knowledge, molecular, and fused features.
        (Right) Boxplot comparison showing that multi-modal fusion features yield the lowest average entropy and variance.
        These results suggest that fusion representations are more compact and potentially more discriminative.}
    \end{subfigure}

    \vspace{1mm}

    \begin{subfigure}[b]{0.9\textwidth}
        \centering
        \includegraphics[width=\textwidth]{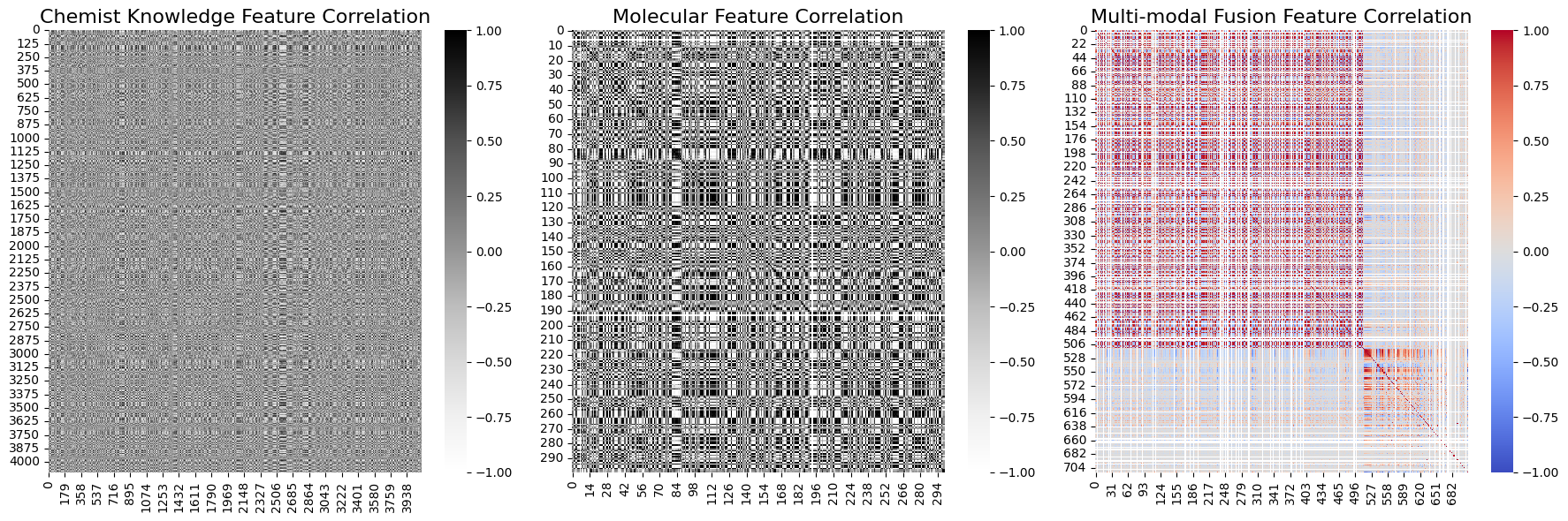}
        \caption{
        Correlation heatmaps of feature representations on the BACE dataset. 
        From left to right: chemist knowledge features, molecular features, and multi-modal fusion features. 
        Compared to the individual modalities, the fused features exhibit stronger and more structured correlations, 
        indicating richer interactions among feature dimensions.
        }
    \end{subfigure}

    \caption{Feature analysis on the BACE dataset comparing multi-modal fusion, chemist knowledge, and molecular features across PCA, entropy, and correlation views.}

    \label{fig:feature_analysis_combined}
\end{figure}

We analyzed the feature space for the BACE dataset using Principal Component Analysis (PCA), Shannon entropy, and correlation heatmaps to understand the internal representations learned by our framework. The comprehensive results are presented in Figure~\ref{fig:feature_analysis_combined}.

Our analysis shows that combining chemist knowledge with molecular structure yields a feature space that is both more informative and more discriminative than either modality alone. In the PCA projections (Figure \ref{fig:feature_analysis_combined}(a)), the fused representations exhibit clearer class separation than the molecular-only or knowledge-only features. Entropy analysis (Figure \ref{fig:feature_analysis_combined}(b)) further indicates that fusion preserves high information content while avoiding excessive redundancy. Finally, correlation heatmaps (Figure \ref{fig:feature_analysis_combined}(c)) reveal new cross-modal dependencies introduced by the fusion step, providing empirical support for our multi-modal design.

\subsection{Feature Contribution Analysis}
\begin{figure}[htbp]
\centering
\begin{subfigure}{0.9\textwidth}
    \includegraphics[width=\textwidth]{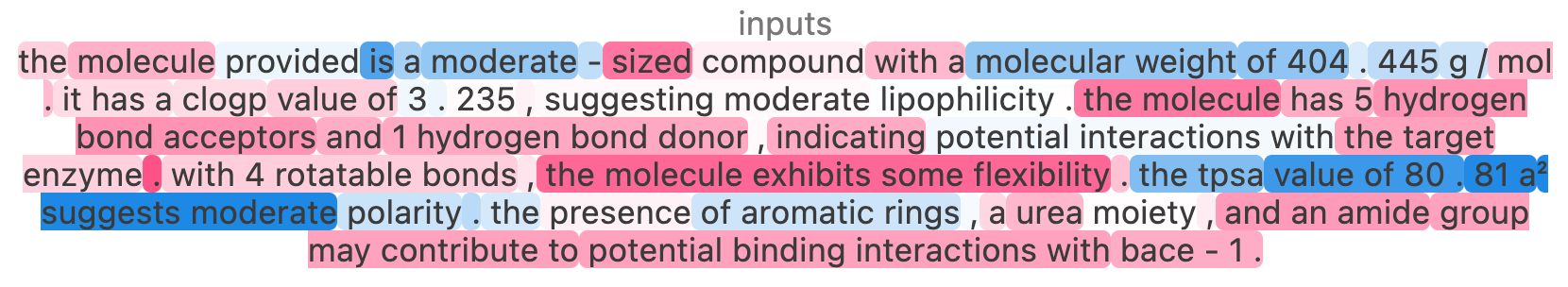}
    \caption{}
    \label{fig:shap1}
\end{subfigure}
\vspace{0.5em}
\begin{subfigure}{0.9\textwidth}
    \includegraphics[width=\textwidth]{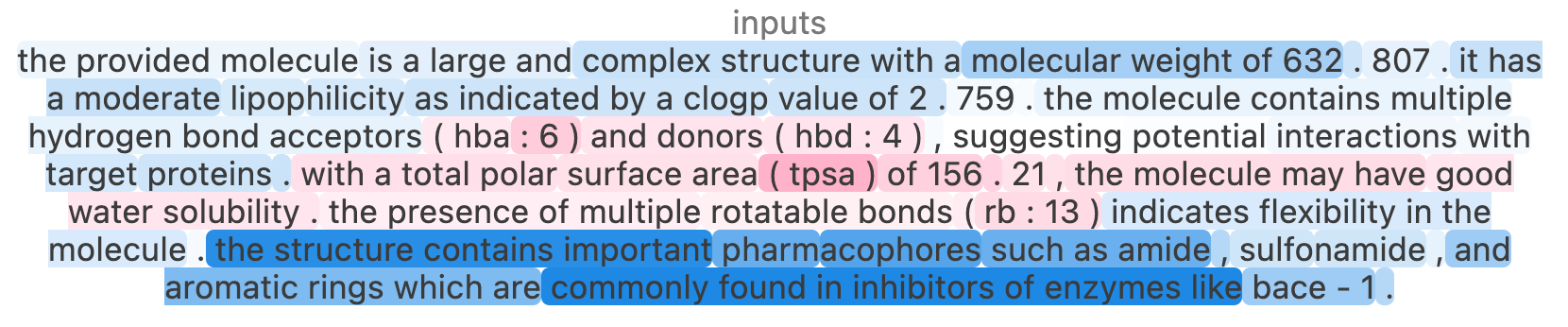}
    \caption{}
    \label{fig:shap2}
\end{subfigure}
\vspace{0.5em}
\begin{subfigure}{0.9\textwidth}
    \includegraphics[width=\textwidth]{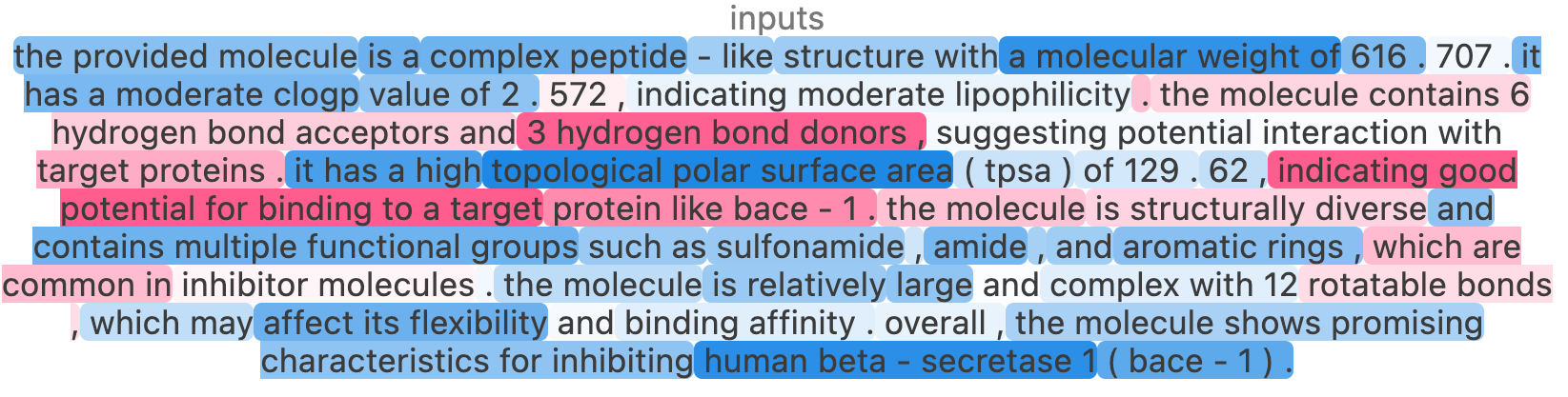}
    \caption{}
    \label{fig:shap3}
\end{subfigure}
\vspace{0.5em}
\begin{subfigure}{0.9\textwidth}
    \includegraphics[width=\textwidth]{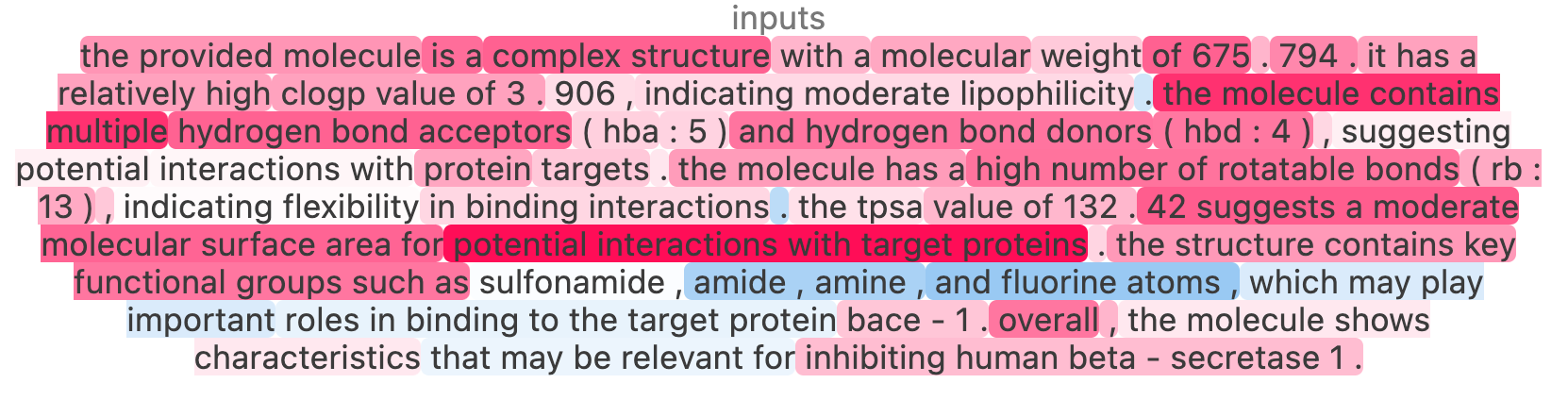}
    \caption{}
    \label{fig:shap4}
\end{subfigure}
\caption{Token-level SHAP visualizations for the MolProphecy model. Blue tokens contribute positively to the predicted class, while red tokens contribute negatively.}
\label{fig:shap_token_importance}
\end{figure}

\begin{figure}[htbp]
\centering
\includegraphics[width=0.95\textwidth]{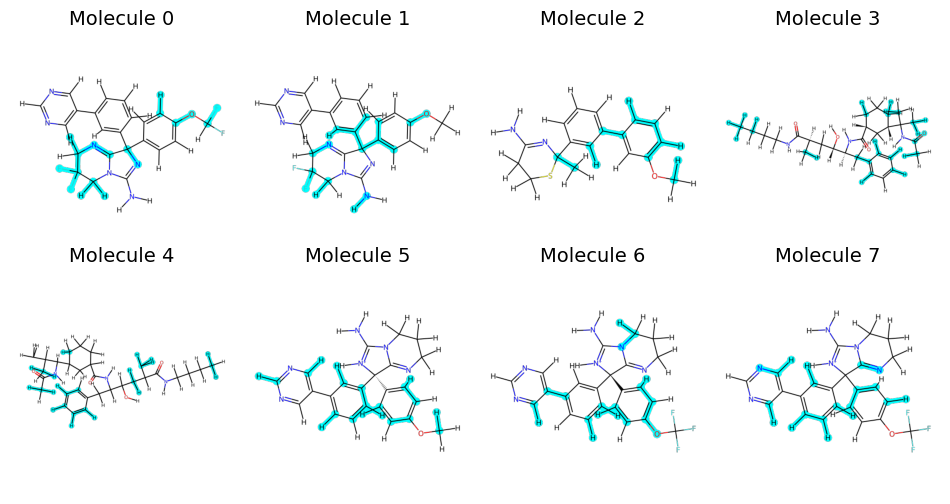}
\caption{GNNExplainer visualizations of four representative BACE inhibitors. Highlighted atoms and bonds indicate substructures most influential to the model's prediction.}
\label{fig:gnn_explain}
\end{figure}

To understand MolProphecy’s decision process, we examined feature attributions from both modalities. SHAP~\citep{lundberg2017unified} yields token-level scores; its visualizations (Figure~\ref{fig:shap_token_importance}) show consistent attention to chemically meaningful terms such as ``molecular weight'' and ``aromatic rings,'' echoing common medicinal-chemistry heuristics. At the structural level, GNNExplainer~\citep{ying2019gnnexplainer} highlights influential subgraphs (Figure~\ref{fig:gnn_explain}), pinpointing key motifs such as amide linkages and polar functional groups that underlie BACE-1 inhibition.

These complementary views confirm that MolProphecy draws on both knowledge and structure. SHAP indicates reliance on high-level concepts (e.g., potency, solubility), whereas GNNExplainer exposes the specific substructures driving activity. Together, they demonstrate improved interpretability that parallels the way medicinal chemists integrate conceptual reasoning with structural detail.

\subsection{Contributions and Limitations}
\ins{MolProphecy introduces a novel HITL framework that tightly integrates chemist-derived domain knowledge with graph-based molecular representations for property prediction. A key distinction from prior work is our treatment of chemist expertise not merely as auxiliary data, but as an independent, co-equal knowledge modality. This expertise, currently simulated by ChatGPT but seamlessly swappable with real chemist input without model retraining, enables a deeper alignment between human reasoning and data-driven modeling. This unique design leads to enhanced predictive performance across both regression and classification tasks on four challenging molecular benchmarks. Furthermore, our proposed gated cross-attention effectively synthesizes structural features with chemist knowledge, improving model robustness, especially under data-scarce and imbalanced scenarios. Interpretability analyses using PCA, entropy, SHAP, and GNNExplainer reveal that the fused representations are more expressive and informative than either modality alone, reinforcing the value of this multi-modal integration.}

\ins{Nonetheless, our framework has several limitations. The reliance on LLM-generated knowledge introduces potential variability, particularly when prompted descriptions are ambiguous or biased. The gated cross-attention layer, while effective, increases model complexity and may limit scalability in resource-constrained settings. Additionally, the current setup assumes static chemist knowledge input per molecule and lacks task-specific prompting or chemist feedback refinement. Future work could address these issues by exploring prompt optimization, adaptive fusion strategies, and interactive HITL workflows to further improve performance and usability.}

\section{Conclusion}

This study presents MolProphecy, a multi-modal framework that integrates chemist-inspired domain knowledge with molecular graph representations to improve molecular property prediction. Extensive experiments demonstrate its superior performance and interpretability across diverse tasks and datasets.

By bridging domain knowledge and data-driven modeling, MolProphecy offers a scalable and interpretable approach toward HITL drug discovery. Future work will explore adaptive prompting, efficiency improvements, and advancements in HITL interaction to further enhance its generalization and usability.

\appendix
\section{Task-Specific Prediction Descriptions}
\label{task_descriptions}

The following descriptions correspond to the \texttt{\{task-specific description\}} fields referenced in the prompt template. They define the prediction objectives for each benchmark dataset used in our experiments.

\begin{itemize}
    \item \textbf{FreeSolv:} This task involves predicting the hydration free energy of small molecules in water, which reflects solvation behavior and plays a key role in drug absorption and distribution.

    \item \textbf{BACE:} A binary classification task for predicting the binding affinity of molecules with human beta-secretase 1 (BACE-1), a therapeutic target in Alzheimer's disease.

    \item \textbf{ClinTox:} A toxicity classification task that aims to distinguish drugs approved by the FDA from those that have failed clinical trials due to toxicity.

    \item \textbf{SIDER:} This multi-label classification task predicts potential adverse drug reactions (ADRs) across 27 physiological systems. The full list of ADR categories is as follows:
    \begin{enumerate}\itemsep0pt
        \item Hepatobiliary disorders
        \item Metabolism and nutrition disorders
        \item Product issues
        \item Eye disorders
        \item Investigations
        \item Musculoskeletal and connective tissue disorders
        \item Gastrointestinal disorders
        \item Social circumstances
        \item Immune system disorders
        \item Reproductive system and breast disorders
        \item Neoplasms (benign, malignant, and unspecified)
        \item General disorders and administration site conditions
        \item Endocrine disorders
        \item Surgical and medical procedures
        \item Vascular disorders
        \item Blood and lymphatic system disorders
        \item Skin and subcutaneous tissue disorders
        \item Congenital, familial, and genetic disorders
        \item Infections and infestations
        \item Respiratory, thoracic, and mediastinal disorders
        \item Psychiatric disorders
        \item Renal and urinary disorders
        \item Pregnancy, puerperium, and perinatal conditions
        \item Ear and labyrinth disorders
        \item Cardiac disorders
        \item Nervous system disorders
        \item Injury, poisoning, and procedural complications
    \end{enumerate}
\end{itemize}

\end{document}